\documentclass[12pt,authoryear]{elsarticle}

\usepackage{natbib}
\usepackage{amssymb}
\usepackage{amsmath,amsfonts}
\usepackage{algorithmic}
\usepackage{algorithm}
\usepackage{array}
\usepackage[caption=false,font=normalsize,labelfont=sf,textfont=sf]{subfig}
\usepackage{textcomp}
\usepackage{stfloats}
\usepackage{url}
\usepackage{verbatim}
\usepackage{graphicx}
\usepackage{booktabs}
\usepackage{bm}
\usepackage{multirow}
\usepackage{soul}
\usepackage{color,xcolor}
\usepackage{hyperref}
\soulregister{\citep}7
\soulregister{\ref}7

\journal{Expert Systems With Applications}

\begin{document}

\begin{frontmatter}



\title{Online Adversarial Knowledge Distillation for Graph Neural Networks}


\author[]{Can Wang}
\ead{wcan@zju.edu.cn}
\author[]{Zhe Wang}
\ead{zhewangcs@zju.edu.cn}
\author[]{Defang Chen}
\ead{defchern@zju.edu.cn}
\author[]{Sheng Zhou\corref{cor1}}
\ead{zhousheng\_zju@zju.edu.cn}
\author[]{Yan Feng}
\ead{fengyan@zju.edu.cn}
\author[]{Chun Chen}
\ead{chenc@zju.edu.cn}

\affiliation[inst1]{organization={College of Computer Science, Zhejiang University; ZJU-Bangsun Joint Research Center; Shanghai Institute for Advanced Study of Zhejiang University.},
            city={Hangzhou},
            postcode={310013}, 
            state={Zhejiang},
            country={China}}
    
\cortext[cor1]{Corresponding author.}

\begin{abstract}
Knowledge distillation, a technique recently gaining popularity for enhancing model generalization in Convolutional Neural Networks (CNNs), operates under the assumption that both teacher and student models are trained on identical data distributions.  
However, its effect on Graph Neural Networks (GNNs) is less than satisfactory since the graph topology and node attributes are prone to evolve, thereby leading to the issue of distribution shift.
In this paper, we tackle this challenge by simultaneously training a group of graph neural networks in an online distillation fashion, where the group knowledge plays a role as a dynamic virtual teacher and the structure changes in graph neural networks are effectively captured. 
To improve the distillation performance, two types of knowledge are transferred among the students to enhance each other: \textit{local knowledge} reflecting information in the graph topology and node attributes, and \textit{global knowledge} reflecting the prediction over classes. We transfer the global knowledge with KL-divergence as the vanilla knowledge distillation does, while exploiting the complicated structure of the local knowledge with an efficient adversarial cyclic learning framework. Extensive experiments verified the effectiveness of our proposed online adversarial distillation approach. The code is published at \url{https://github.com/wangz3066/OnlineDistillGCN}.
\end{abstract}


\begin{highlights}
\item We propose an Online Adversarial Knowledge Distillation framework for GNNs.
\item OAD trains a group of GNN models to capture structure updates in GNNs.
\item OAD extracts both local and global knowledge in graph structured data.
\item Extensive experiments demonstrate the effectiveness of OAD.

\end{highlights}


\end{frontmatter}

\section{Introduction}
Recent advances in computing devices such as graphical processing units are rendering possible to train Deep Neural Networks (DNNs) with millions of parameters. In particular, Deep Convolutional Neural Networks (DCNNs) have made dramatic breakthroughs in a variety of computer vision applications including image classification \citep{simonyan2015very} and object detection \citep{he2017mask}, etc, due to its extraordinary ability on feature extraction and expression. However, these DCNNs often contain tens or even hundreds of layers with millions of trainable parameters. For example, Inception V3 network \citep{szegedy2015going} requires about 24M trainable parameters while ResNet152 \citep{he2016deep} contains about 60M trainable parameters. Training such cumbersome models is highly time-consuming and space-demanding, thus their deployment on resource-limited devices such as mobile phones or tablets is restricted. For years, many approaches have been proposed to transform the heavy neural network into a lightweight one, such as network pruning \citep{wu2016quantized} and low-rank factorization \citep{denton2014exploiting}.  Knowledge Distillation (KD) is also proposed to compress and accelerate those cumbersome \textit{teacher} models by training a lightweight \textit{student} model to align with the teacher' predictions \citep{hinton2015distilling}. The predictions from teacher are used as a positive regularization to improve the generalization ability of the student network. Besides prediction alignment, knowledge distillation could also be achieved by aligning intermediate representations \citep{romero2014fitnets,chen2021cross} and sample relations \citep{passalis18learning,park19relational} between the teacher and student models. Due to its simple implementation and effectiveness, knowledge distillation has gained much popularity.

Recently, knowledge distillation has attracted the attention of the Graph Neural Networks (GNNs) community, where a tiny student GNN model is trained by distilling knowledge from a large pre-trained teacher GNN model \citep{yang2020distilling,yang2021extract}. A successful knowledge distillation requires that the data distribution of both the teacher and student models are similar. Nevertheless, directly employing the vanilla teacher-student paradigm will be plagued by the volatility of the graph data. That is, the graph topology and node attributes are likely to change over time. For example, in social networks, new nodes or new edges may be added to the graph. This indicates the distributions of training sets for the teacher and student models might vary. Consequently, the student model trained with outdated knowledge from the static teacher will probably result in sub-optimal solutions. A static teacher model therefore is not optimal for transferring knowledge in a volatile environment. Although this distribution shift problem has been studied for Convolutional Neural Networks (CNNs) \citep{nguyen2021knowledge}, the problem in the graph domain has not been well addressed. The challenge of this problem lies in the constant evolution of graph data. Traditional distribution alignment techniques, such as transfer learning, would be ineffective because they assume that the data distributions from the source and target domains are fixed. Besides, pretraining a large teacher model requires significantly more processing time and computation resources than training a student model, resulting in many economic and environmental problems. Considering the expensive cost and the frequency of the change in graph-structured data training, retraining the heavy-weight teacher GNN when the graph changes will be impractical.  
To alleviate this problem, instead of training a single heavy-weight static teacher GNN, we simultaneously train a group of student GNNs in an online fashion to effectively capture dynamic changes and leverage the group knowledge as an effective proxy for the pre-trained teacher model. 
In this way, each student GNN will distill the knowledge from other peers to enhance its performance. When retraining is required in case of graph updates, much less cost will be incurred by the light-weight student models than the heavy-weight teacher models.

More specifically, each student GNN is trained with the conventional cross-entropy loss from the ground truth labels as well as two types of knowledge from other peers, which corresponds to two characteristics of graph data: \textit{Local Knowledge} and \textit{Global Knowledge}: 
(1) \textit{Local Knowledge} reflects information contained in the graph topology and node attributes, which is instantiated as the learned node embedding with aggregated attributes from neighboring nodes. To better capture the complicated structure of the local knowledge, we employ the adversarial cyclic learning to effectively achieve the embedding alignment among students.
(2) \textit{Global Knowledge} reflects the prediction over classes by the semi-supervised classifiers. We transfer this type of knowledge with KL divergence as the vanilla knowledge distillation does.

We conduct extensive experiments on benchmark datasets with different GNN architectures to verify the generalization of our proposed \textbf{O}nline \textbf{A}dversarial knowledge \textbf{D}istillation (OAD) framework. Ablation studies are also designed to demonstrate the necessity of each proposed component.

In summary, the contributions of this paper are concluded as follows:
\begin{itemize}
\item  A novel Online Adversarial Distillation (OAD) framework is proposed to simultaneously train a group of student GNN models in an online fashion in which they can learn from peers and  effectively capture structure updates in graph neural networks.
\item Student GNNs are trained with both global and local knowledge in the GNNs group for better distillation performance. To learn the complicated structure of the local knowledge, adversarial cyclic learning is employed to achieve more accurate embedding alignment among students.
\item Extensive experiments including transductive/inductive node classification and object recognition on benchmark datasets as well as dynamic graphs demonstrate the effectiveness of our framework.
\end{itemize}

The structure of this paper is organized as follows. In Section \ref{Related works}, we review the literatures that closely related to the topic of this paper. In Section \ref{method}, we introduce the details of the proposed Online Adversarial knowledge Distillation (OAD) for GNNs. In Section \ref{experiment}, we compare the performance of proposed OAD model with other baseline methods to demonstrate its effectiveness. The conclusions and future works of this paper is presented in Section \ref{conclusion}.

\section{Related works}
\label{Related works}
\subsection{Knowledge distillation} 

Knowledge distillation is a model compression technique which aims to boost the performance of a lightweight student model leveraging the 'dark knowledge' of a over-parameterized teacher model. 
The first work of knowledge distillation is introduced by \citep{hinton2015distilling}, which regards the prediction of teacher model as knowledge and transfers it to student model using Kullback-Leibler (KL) divergence.
To further enhance the performance, subsequent works attempted to align feature maps from intermediate layers \citep{romero2014fitnets, zagoruyko2016paying,chen2021cross,sepahvand2022teacher} or align sample relations based on their representations from the penultimate layer \citep{passalis18learning,park19relational,chen2022simkd}. 
To circumvent the pretraining step for obtaining a large teacher model, online knowledge distillation provides a more economic solution by simultaneously training a group of student models and encouraging each student to distill from other peers \citep{zhang2018deep,lan2018knowledge,chen2020online, zhang2022exploring}. The reason behind these works is that we can dynamically construct one or several ``virtual'' teacher(s) with higher accuracy to guide the student during training. The most straightforward way to obtain the ``virtual'' teacher(s) is to simply average the predictions of other peers \citep{zhang2018deep}, which is later improved by adaptively learning the weights through gate or self-attention mechanism \citep{lan2018knowledge,chen2020online}. Although online knowledge distillation has been extensively studied on CNNs, to the best of our knowledge, our work is the first attempt on GNNs. 

\subsection{Graph Neural Networks}
Graph learning has many useful real-world applications such as hyperspectral image classification \citep{hong2020graph}, remote sensing \citep{hong2019learnable, hong2019learning}. Recently, deep learning based model such as Graph Neural Networks have dominated many graph learning tasks. Early works of GNN are mostly spectral based, which firstly transforms graph signal into spectral domain via graph Fourier Transformation and conducts convolution on the spectral domain \citep{bruna2014spectral, henaff2015deep}. However, the computation complexity of spectral GNN is $O(N^3)$ due to eigen decomposition of Laplacian matrix, which is time-consuming. ChebyNet \citep{defferrard2016convolutional} simplifies spectral GNN via K-order polynomial approximation of the graph Laplacian matrix. Graph Convolutional Network (GCN) \citep{kipf2017semi} further introduces the first-order approximation of ChebyNet, which can also be viewed as the simplest spatial-based GNN. Spatial-based GNN uses a message passing process by aggregating the messages of neighbors recursively, thus the information from higher-order neighbors will be passed to lower-order neighbors. Graph Attention Network (GAT) introduced self-attention mechanism to allow more flexible importance assignment to neighbors \citep{velivckovic2018graph}. GraphSAGE generated embeddings by sampling and aggregating neighbor features, enabling inductive learning on large-scale graphs \citep{hamilton2017inductive}. Graph Isomorphism Network (GIN) \citep{xu2018powerful} proved that the message-passing GNN is at most as powerful as 1-Weisfeiler Lehman Test, based on which they design the most powerful GNN that can distinguish any non-isomorphic neighborhoods. Some other GNNs are designed for 3D point cloud recognition tasks where the graphs are generated with k nearest neighbors \citep{landrieu2018large,wang2019dynamic}.

Although the GNN is effective in handling graph data, it still suffers from the problem caused by redundant network parameters. This problem has been addressed by some network compression techniques, such as pruning \citep{chen2021unified}. Some works also explored the potential of knowledge distillation to improve the performance of GNNs \citep{yang2020distilling, yang2021extract}. The student models of \citep{yang2021extract} are parameterized label propagation module instead of GNNs. LSP \citep{yang2020distilling} transferred local structure knowledge of a teacher GNN to a student GNN with less parameters. The main differences between the proposed OAD and existing knowledge distillation with GNN methods are two-fold: Firstly, existing knowledge distillation with GNN methods are mostly offline, where a teacher model is pretrained and a student model is trained to distill the knowledge of teacher. Since the graph-structured data changes frequently, the data distributions of teacher and student may vary. Therefore, the knowledge from the pretrained teacher GNN may be ineffective for the student GNN. Our method uses an online knowledge distillation framework for GNNs, where multiple student models are trained simultaneously, which ensures that the data distribution of students and "virtual teacher" is always aligned. Secondly, existing knowledge distillation with GNN methods such as LSP \citep{yang2020distilling} mostly align the intermediate feature maps via hand-crafted distance measure such as L2 distance or KL divergence. However, many hand-crafted distance measures are not defined in highly correlated intermediate features \citep{tian2020contrastive}, thus they may be ineffective in transferring the knowledge of intermediate layers. Instead, in our proposed OAD framework,  we employ a cyclic GAN framework for local knowledge distillation, where some discriminators are employed to implicitly supervise the distribution alignment of teacher and student model. Therefore, the intermediate feature of students and virtual teachers are automatically aligned under the supervision of discriminators.

\subsection{Generative Adversarial Networks} 

Generative Adversarial Network (GAN) is a generative model initially proposed by \citep{goodfellow2014generative}. It is consisted by a generator and a discriminator, where the generator attempts to synthesis new data samples similar to original dataset, and the discriminator is adopted to evaluate the authenticity of data instances. Due to its strong ability to generate high-quality samples, GANs have received significant attention. GAN has been improved from different aspects. To better control the mode of generated data, some works proposed to condition GAN on the latent variables \citep{mirza2014conditional,chen2016infogan}. Some works adopt more elaborate distance metrics to increase the training stability \citep{gulrajani2017improved,mao2017least}. Recently, in the KD field, the discriminator is included to enforce the student network to mimic the teacher network's pattern \citep{wang2018kdgan,liu2018teacher}. Some existing works have employed GAN to generate new graph with node features and adjacency matrices \citep{de2018molgan,bojchevski2018netgan} similar to original graphs. Our works are significantly different from aforementioned works. Specifically, our OAD leverages GAN to enable local knowledge transfer among different student GNNs while these works use GAN to generate new data samples.

\begin{figure*}
\centering
\includegraphics[width=\textwidth]{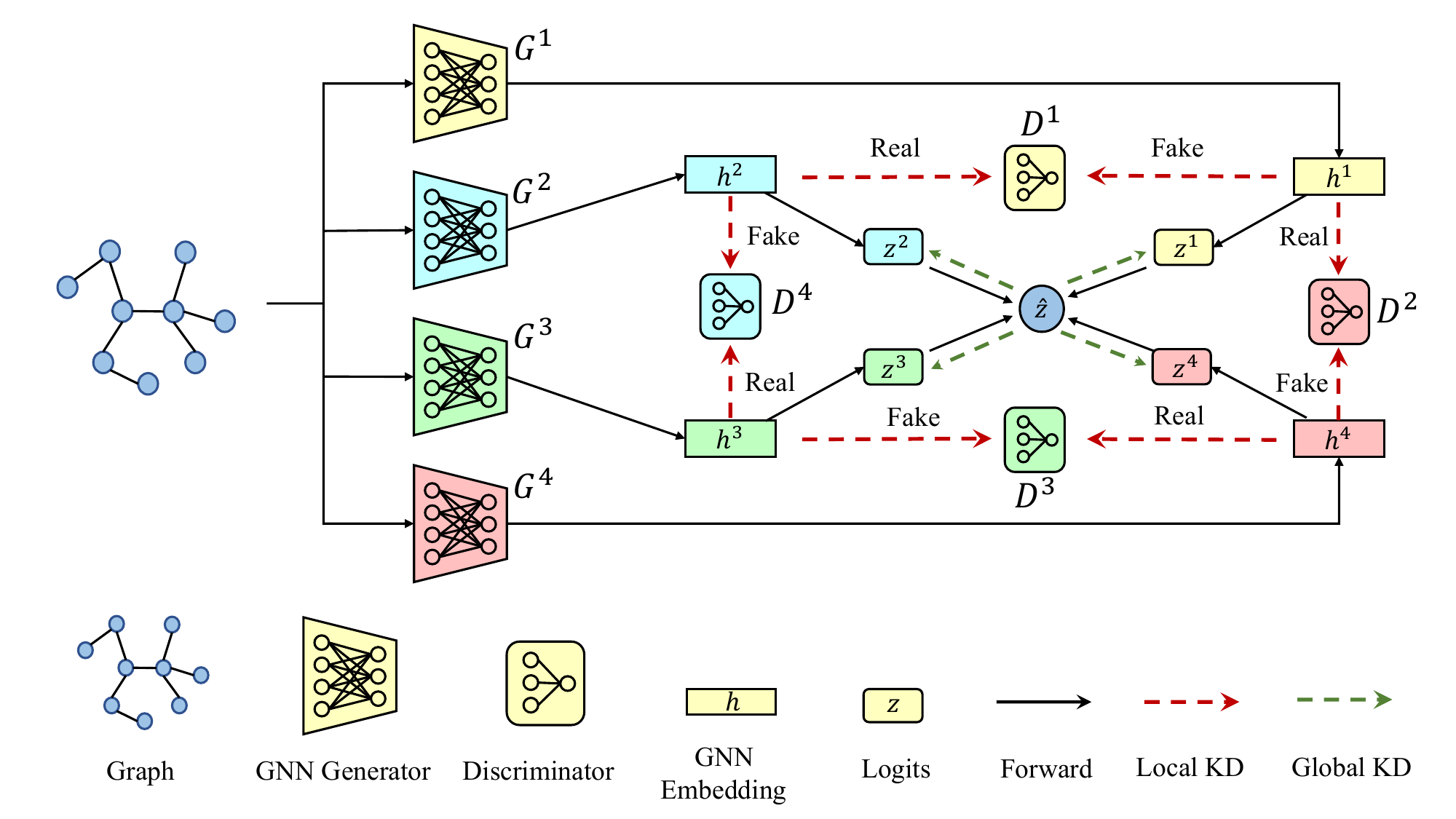}
\caption{Overview of the proposed online adversarial knowledge distillation framework for GNNs. We show the case with four student GNN models. The student network is a GNN generator. Each student model is assigned with a discriminator. The local knowledge is distilled by a cyclic generator-discriminator framework, denoted as red dot lines. The global knowledge is distilled by the KL divergence between student logits and "virtual" teacher's logits, denoted as green dot lines. }
\label{fig:model}
\end{figure*}

\section{Method}
\label{method}
In this section, we will elaborate on our proposed \textbf{O}nline \textbf{A}dversarial knowledge \textbf{D}istillation (OAD) framework for graph neural networks.
Generally, we will utilize the online knowledge distillation paradigm to simultaneously train a group of student models and encourage each student to learn from other peers mutually.

More specifically, we start with a group of untrained student models with randomly initialized parameters. 
Intuitively, these students are expected to capture different characteristics of graph data, which can be mainly grouped into two types:
(1) Local knowledge combining the local neighborhood structure and the node features, which is naturally represented by the learned node embeddings.
(2) Global knowledge referring to the label distribution predicted by the GNN models, deciding the corresponding categories for each node.

Both of these two knowledge types are transferred among students while they differ in the way of distillation.
As the global knowledge is represented by the distribution over the labels, we follow the vanilla knowledge distillation and use the Kullback-Leibler divergence (KL) to measure the distance. In contrast, the intermediate layers of GNN models are formulated by aggregating the messages from K-hop neighbors, which encodes rich local subgraph information and is more complicated than the label distribution. How to ditill the knowledge of intermediate layers can be referred as a "local knowledge distillation" process. More importantly, each student model is expected to capture different characteristics of the graph. Strictly aligning the embedding with $L_{2}$ or KL measurement will impair the diversity and degenerates the knowledge transferred among student models. Inspired by the success of adversarial training in learning complicated distribution, we train several discriminators to distinguish the difference between the learned embedding distribution, which yields more accurate results.
Additionally, we utilize the cyclic training strategy to improve the training efficiency. Figure \ref{fig:model} illustrates the proposed OAD framework, which contains three major components: local knowledge distillation, global knowledge distillation, and adversarial cyclic training. 
In the following contents, we will first recap some fundamental concepts about knowledge distillation and graph neural networks. Then, we introduce the details about the proposed model. 

\subsection{Supervised Learning}
    In a supervised C-class classification problem, we are given a dataset with $m$ labeled samples $\mathcal{D}=\{\bm{x}^{(i)}, \bm{y}^{(i)}\}_{i=1}^m$, where $\bm{x}^{(i)} \in \mathcal{R}^d$ is the input feature sampled from an unknown data distribution $\mathcal{X}$ and $\bm{y}^{(i)} \in \mathcal{R}^{C}$ is the one-hot ground truth label. Our goal is to learn a classifier parameterized by $\Theta$: $f(\cdot; \Theta) : \mathcal{X} \rightarrow [0,1]^C$. The output of mapping function $f$ is known as "logits" $\bm{z}$, which is subsequently transferred into probability distribution $\bm{p}=\sigma(\bm{z}) \in \mathcal{R}^C$ with a softmax function $\sigma(\cdot)$ :
    \begin{equation}
        p_i = \dfrac{\exp(z_i)}{\sum_{j=1}^C \exp(z_j)}
    \end{equation}
    The classifier is typically trained to minimize the cross entropy (CE) between probability distribution $\bm{p}$ and ground truth labels $\bm{y}$ :
    \begin{equation}
        \mathcal{L}_{CE}(\bm{p}, \bm{y}) = \sum_{i=1}^C y_i \log p_i
    \end{equation}

\subsection{Knowledge Distillation}
   Knowledge distillation is generally composed by two neural networks with distinct architectures: A teacher network with larger parameters and a student network with fewer parameters. Besides standard cross entropy loss between student predictions \bm{$p_s$} and ground truth label \bm{$y$}, student network is also trained to minimize the Kullback-Leibler (KL) divergence between softened student probability $\sigma(\bm{z_s}/T)$ and softened teacher probability $\sigma(\bm{z_t}/T)$: 
    \begin{equation}
        \mathcal{L}_{KD} = \mathcal{L}_{CE}(\bm{p}_s,\bm{y})+\alpha T^2KL(\sigma(\bm{z}_s/T),\sigma(\bm{z}_t/T))
    \end{equation}
    where $T$ is a hyper-parameter known as temperature. The higher $T$ leads to more significant softened effect.  
    
    Besides output logits, the teacher networks' intermediate layers also encoded highly representative knowledge to the given dataset, which may be potential to boost the students performance further. Many works have been proposed to match the raw or transformed hidden layers. For example, Fitnet \citep{romero2014fitnets} randomly selects a hidden layer from both teacher network and student network and minimize their mean square error (MSE) loss:
    \begin{equation}
        \mathcal{L}_{Fitnet} = \dfrac{1}{2}|| r(\bm{h}_s)-\bm{h}_t||^2 
    \end{equation}
        where $r(\cdot)$ is a dimension transformation operator such as Multi-Layer Perception.  

\subsection{Graph Neural Networks}
In this subsection, we briefly recall the definition of GNNs. The input of a GNN is a graph denoted as $ \mathcal{G}=\{\mathcal{V},\mathcal{E}\} $ where $\mathcal{V}=\{v_1, v_2, ...,v_N\} $ is the node set and $E$ is the edge set. 
The node features are given by a matrix $X=\{\bm{x}_1, \bm{x}_2, ...,\bm{x}_N\} \in \mathcal{R}^{N \times D} $, where $\bm{x}_i \in \mathcal{R}^D $ is the node feature of $v_i$. 
The adjacency matrix of $G$ is denoted by $A^{N \times N}$ where $A_{ij}=1$ if $(v_i,v_j) \in \mathcal{E}$ and $A_{ij}=0$ if $(v_i,v_j) \notin \mathcal{E}$. 
The connected neighborhood set of node $v_{i}$ is denoted as $\mathcal{N}_{i}$.

Graph neural networks aim at learning a mapping function $f: \mathcal{R}^{N \times N} \times \mathcal{R}^{N \times D} \rightarrow \mathcal{R}^{N \times D'}$.
The core operation of graph neural networks is aggregating features from the neighborhood nodes.
A typical GNN with $L$ layers is defined as:
\begin{equation}
\begin{aligned}
   & G(\bm{x}_{i}) = \bm{h}^{(L)}_{i}, \quad \bm{h}^{(l+1)}_i= \psi(\bm{h}^{(l)}_i, \mathcal{AGG}_{j\in \mathcal{N}_{i}} \phi(\bm{h}^{(l)}_i, \bm{h}^{(l)}_j))
\end{aligned}
\end{equation}
where $l=1,2,...,L-1$, $\bm{h}_i^{(l)}$ is the embedding of node $v_{i}$ in the $l$-th hidden layer of the GNN encoder $G$, $\phi(\cdot,\cdot)$ is a pairwise message passing function, $\mathcal{AGG}$ function is used to aggregate messages from all neighbors of $v_i$, and $\psi$ function updates the state of $v_i$ based on its current state as well as the messages propagated by local neighbors. 
Existing GNN methods have mainly varies on the message passing functions $\phi(\cdot,\cdot)$ as well as the aggregation function $\mathcal{AGG}$.
To verify the generalization of our proposed framework for the graph neural networks, we test three widely adopted graph neural networks named GCN\citep{kipf2017semi}, GAT\citep{velivckovic2018graph} and GraphSAGE\citep{hamilton2017inductive}.

\subsection{Local Knowledge Distillation}
The aforementioned definition of GNNs demonstrates that the learned embedding $\bm{h}_{i}$ can effectively represent the local information around node $v_{i}$. 
Inspired by the idea of intermediate-layer association in knowledge distillation, a straightforward strategy to enhance representation ability of $\bm{h}_i$ is to distill the intermediate layer or its transformation from a better generalized teacher GNN.
However, we propose that this method suffers from following two drawbacks: 1) It requires a teacher GNN usually with much more parameters being pretrained, which needs expensive computations. 2) More importantly, the graph-structured data can be more frequently changed than static data such as images. For example, in a social network, there are millions of new following relationships generated everyday. Thus, the dataset distributions of teacher and student network may vary given the drastic changes of graph, and the knowledge learned by teacher network may not be helpful for the student network. The experiments on dynamic graph prove this point, see subsection \ref{dynamic graph}. 

Instead, we propose Online Adversarial Distillation (OAD) for GNN to address above two drawbacks.
In OAD framework, a group of lightweight student GNNs are trained simutanesouly and each student is expected to distill the knowledge from a "virtual" teacher formulated by the remaining group members, thus a pretrained teacher network is not needed.
Since the total number of parameters in OAD framework is less than pretrained teacher network, it is more spatially efficient than vanilla knowledge distillation framework. 
Besides, as all the networks are trained on the same graph, the problem of domain shift caused by the dynamicity of the graph is solved. 

Next, we introduce the local knowledge distillation module of OAD. 
The $m$-th student GNN is expected to distill the knowledge from the intermediate layers of the remaining $M-1$ GNNs.
We adopt a GAN framework to facility the distillation procedure.
For the $m$-th student network, its graph convolution layers $G^{m}$ are utilized as a generator to generate embedding $\bm{h}^m$.
Then, to distill the knowledge of embedding $\bm{h}_q$ generated by another GNN members $G^q$, a discriminator $D^{mq}:\mathcal{R}^{N \times D'} \rightarrow \mathcal{R}^N $ is included. 
The discriminator inputs $h^m$ or $h^q$ and outputs a scalar between 0 (fake) and 1 (real).
When the input is $\bm{h}^m$, $D^{mq}$ is trained to output 0 (fake).
When the input is $\bm{h}^q$, $D^{mq}$ is trained to output 1 (real).
The GNN generator $G^m $ is trained to confuse the discriminator and obtain high score (close to real) after discrimination. 

However, discriminating all pairs of $M$ student models takes $M*(M-1)/2$ times, which is unfriendly to efficient training, especially when the number of student model grows.
To improve the training efficiency while maintaining the performance, we follow \citep{chung2020feature} and utilize the cyclic training strategy to reduce the number of discrimination into $M$.
For $D^m$, it takes $G^m$ and $G^{m+1}$ as input and learns to discriminate $G^m$ as fake and $G^{m+1}$ as real. 
In this way, each student GNN generator is trained to confuse the $D^m$ and generate the embedding that follows the distribution of other student models.
For the entire students group, the knowledge of node embeddings are transferred in a cycle: $M \rightarrow M-1$, $M-1 \rightarrow M-2$, ..., $1 \rightarrow M$.
We use the original GAN framework where both discriminators and generators are trained to minimize the following loss:
\begin{equation}
\begin{split}
& \mathcal{L}_{D} = \sum_{m=1}^{M-1} [\log D^m(G^m(X))-\log D^m(G^{m+1}(X))]  \\ &  +\log D^M(G^M(X))-\log D^M(G^{1}(X)) \\
& \mathcal{L}_{G} = \sum_{m=1}^M - \log D^m(G^m(X))
\end{split}
\end{equation}
In the training phase, sequentially updating discriminator $D^{m}$ and GNN encoder $G^{m}$ will destroy the computation graph since $G^{m}$ will also join the loss calculation of $D^{m+1}$. Instead, we fix generators and update all the discriminators firstly, then we fix discriminators and update all the generators. This group optimization process will not increase the training burden.

\subsection{Global Knowledge Distillation}
Given the embedding $\bm{h}^{m}_{i}$ of node $v_{i}$ learned by the $m$-th GNN generator, we use the graph convolutional layer $\Theta$ and softmax function to transform the representation into the logits $\bm{z}^{m}_{i}$, which is computed as:
\begin{equation}
    z^{mk}_{i} =\frac{exp(\Theta(h^{mk}_{i})/\mathcal{T})}{\sum^{K}_{j=1}exp(\Theta(h^{mj}_{i})/\mathcal{T})}
\end{equation}
where $z^{mk}_{i}$ is the probability of node $v_{i}$ in class $k$ predicted by $m$-th student model, $\mathcal{T}$ is the temperature. Note that $\mathcal{T}$ is greater than 1 and we set $\mathcal{T}=3.0$ for all the experiments in this paper.

Note that network parameters of $M$ students are initialized differently, thus their logits are different during the training process. For the $m$-th student GNN, the remaining M-1 different students can act as an "ensemble" teacher, which is generally more effective than single model. We verify this point in Section \ref{subsection:benefit of mutual learning}. Therefore, In the global knowledge distillation, for the $m$-th student GNN model, we formulate the global knowledge of this "ensemble" teacher by averaging their logits $\bm{z}_i^j$ as:  
\begin{equation}
    \widehat{\bm{z}^{m}_{i}} = \frac{1}{M-1}\sum^{M}_{j,j\neq m}\bm{z}^{j}_{i}
\end{equation}

We use the KL divergence between these two softened class distribution as the global knowledge distillation loss:
\begin{equation}
    \mathcal{L}_{B}=\mathcal{T}^2\sum^{N}_{i=1}\sum_{m=1}^M KL(\bm{z}^{m}_{i},\widehat{\bm{z}^{m}_{i}})
\end{equation}
where $\mathcal{T}^2$ is multiplied to keep the gradient this loss term roughly unchanged when $\mathcal{T}$ changes \citep{hinton2015distilling}.

\subsection{Overall Training Process}
The overall training loss for student GNN models is formulated as:
\begin{equation}
    \mathcal{L}_{total} = \mathcal{L}_{CE}+\alpha \mathcal{L}_{G} + \beta \mathcal{L}_{B}
\end{equation}
where $\mathcal{L}_{CE}$ is the supervised cross entropy loss, $\alpha$ and $\beta$ are the coefficients for the corresponding distillation losses. 
In our experiments, each student do not distill the knowledge from other peers at the initial epochs of training, i.e., they only learn from the ground-truth labels. After warmed-up phase, each student has its own perception to the mapping structure, then the meaningful knowledge can be distilled to others. 

\subsection{Complexity Analysis}

In this subsection, we analyze the computation complexity of the proposed OAD framework. Let $T_G^i$ and $T_\Theta^i$ be the computing time of one forward propagation on graph encoder and output layers of the $i$-th student, respectively. Let $T_D^i$ be the computing time of one forward propagation on the $i$-th discriminator. The overall computing time of one forward propagation of proposed model is:
\begin{equation}
\begin{split}
    T_{OAD} &= \sum_{i=1}^M T_G^i + \sum_{i=1}^M T_\Theta^i + \sum_{i=1}^M 2T_D^i \\
      &=  O(\sum_{i=1}^M T_G^i) + O(\sum_{i=1}^M T_\Theta^i) + O(\sum_{i=1}^M T_D^i)
\end{split}
\end{equation}

Thus, the computing time of the proposed OAD model increases linearly w.r.t. the number of group members. Meanwhile, let $N_G$ and $N_\Theta$ be the parameter number of graph encoder and output layers for a single student GNN, respectively. Let $N_D$ be the parameter number of a single discriminator. Thus the total parameter number of OAD model is:
\begin{equation}
    S_{OAD} = M(N_G+N_\Theta+N_D)
\end{equation}

when $M$ is limited, the total number of parameters is much less than a cumbersome GNN.
    
\begin{algorithm}[!h]
    \label{algorithm1}
	\caption{Online adversarial distillation for Graph Neural Network}
	\renewcommand{\algorithmicrequire}{\textbf{Input:}}
    \renewcommand{\algorithmicensure}{\textbf{Output:}}
	\begin{algorithmic}[1]
        \REQUIRE Graph $\mathcal{G}=\{\mathcal{V},\mathcal{E}\}$. Node attribution matrix $X=\{x_1, x_2, ..., x_N\} \in \mathcal{R}^{N \times D}$. Adjacency matrix $A \in \mathcal{R}^{N \times N}$. $M$ student graph neural networks $\{f_m\}_{m=1}^M$. $M$ discriminators $\{D_m\}_{m=1}^M$. Epochs of independent learning $epoch_t$, epochs of online knowledge distillation $epoch_o$, weights of loss $\alpha, \beta$.
        \ENSURE Well-trained graph neural network ensemble.
        \FOR{$t=1$ to $epoch_t$}
            \STATE{Forward $M$ student GNN models to generate predictions $\{f_m(X)\}_{m=1}^M$.}
            \STATE{Compute cross entropy loss $\mathcal{L}_{CE}$ for each GNN.}
            \STATE{Update the parameters of $M$ students by backward propagation of $\mathcal{L}_{CE}$.}
        \ENDFOR
        \FOR{$t=1$ to $epoch_o$}
            \STATE{Forward $M$ student GNN models to generate predictions $\{f_m(X)\}_{m=1}^M$ and hidden embedding $\{G_m(X)\}_{m=1}^M$.}
            \STATE{Forward $M$ discriminators $\{D_m\}_{m=1}^M$.}
            \STATE{Compute $\mathcal{L}_D$.}
            \STATE{Update the parameters of $M$ discriminators by back propagating $\mathcal{L}_D$.}
            \STATE{Compute cross entropy loss $\mathcal{L}_{CE}$ for each student GNN.}
            \STATE{Compute global knowledge distillation loss $\mathcal{L}_B$.}
            \STATE{Compute local knowledge distillation loss $\mathcal{L}_G$.}
            \STATE{Update the parameters of $M$ student GNNs by back propagation of $\mathcal{L}_{total}$.}
        \ENDFOR
	\end{algorithmic}
\end{algorithm}

\section{Experiment}
\label{experiment}
To evaluate the performance of proposed OAD framework, we conduct node classification tasks on citation datasets, Protein-Protein Interaction (PPI) dataset, and object recognition task on ModelNet40 dataset. 
A variety of classic GNN architectures are tested to evaluate the generalization of the OAD framework, including GCN\citep{kipf2017semi}, GAT\citep{velivckovic2018graph} and GraphSAGE\citep{hamilton2017inductive}.
Besides, we compare the performance of OAD and vanilla KD method on dynamic graphs. The result indicates that proposed OAD method is more robust to the drastic changes of graphs than vanilla KD method.
We also conduct an ablation study to evaluate the effectiveness of each module and the group mutual learning strategy in the OAD framework. 
Furthermore, we analyze the impact of group size used in online knowledge distillation to the performance of model.
Finally, we visualize the feature space generated by different models, which demonstrates that proposed OAD framework can generate much more effective representations.
The proposed OAD model is implemented using Pytorch-1.9.1 Library. 
All the experiments are conducted on a Linux Ubuntu 18.04 Server with one GeForce RTX 2080 Ti GPU of 11.0 GB memory.

\subsection{Transductive node classification on citation datasets}
\label{subsection:citation}
We firstly evaluate our model by transductive node classification task on three citation datasets: Cora, Pubmed and Citeceer. 
In these datasets, nodes represent the publications and edges represent citations between them. 
The Cora dataset contains 2708 nodes, 5429 edges, 1433 features and seven possible classes. 
The Pubmed dataset has 19717 nodes, 44338 edges, 500 features and three possible classes. 
The Citeceer dataset contains 3327 nodes, 4732 edges, 3703 edges and six possible classes. 
Following the same experimental settings of existing methods \citep{yang2016revisiting}, 20 nodes per class are selected for training, 500 nodes are for validation and 1000 nodes are for testing. 
The nodes in all of these datasets have only one label, so we use Top-1 Accuracy as the evaluation metric.
In the transductive training setting, the model can access all nodes on the whole graph.  

We compare the performance of vanilla GNNs and the corresponding OAD variants, denoted as ``a specific model + OAD''. 
Each student GNN with the OAD framework has the same network architecture as its corresponding vanilla GNN. 
The results of OAD are the average performance of the student models in the group.
The results of '' A specific model + KD '' are obtained as follows: we firstly pre-train a strong enough teacher model as save its parameters. Then, we use this teacher model the supervise the training of a small-size student model, along with the supervision of ground truth label.

We also include the performance of a GNN ensemble with the same group size and network architecture as OAD framework.
All the hyperparameters settings of "Ensemble" are same as "single GNN" except different group size.  
Note that at inference phase, \textit{an ensemble model averages all the students' predictions while the output of proposed OAD method is the prediction of single student GNN.}

Table \ref{citation architecture} summarizes the network architectures and their parameter numbers in citation experiment. We set the group size in this experiment as 4. As can be seen from Table \ref{citation architecture}, \textit{the total parameters of student networks and extern discrminators are less than one teacher network}.  We adopt two-layer GNNs in all the experiments. For GCN and GraphSAGE, the dimension of hidden layer is 16. For GAT, following the settings in original paper \citep{velivckovic2018graph}, the hidden dimension is 8 and the attention heads are set as [8,1] for Cora and Citeceer, and [8,8] for Pubmed, respectively. The aggregation type of GraphSAGE is "mean". For all the experiments, we fix learning rate, weight decay and optimizer as 0.005, 0.0005 and Adam, respectively. The OAD model is warmed up for 100 epoch, and further mutual learned for 100 epoch. For other baselines, the total traning epoch is 200. For OAD, we adopt GCNs as skeleton of discriminators. The input of discriminator is the last hidden layers (the layer before last convolution layer) of GNNs. The discriminators 
used in GCN experiments are a single-layer GCN, while the discriminators used in GAT and GraphSAGE experiments are two-layer GCNs with hidden dimensions of 16. We set the loss coefficient $\alpha=1$ and $\beta=1$ for all the network-dataset combinations. Note that we evaluate the performance of all the student models and report the best case for our model.

We compare the classification accuracy of different methods. The accuracy is defined as the proportion of number of correctly predicted samples to the number of all samples.
Tabel \ref{citation result} illustrates the experimental results of the transductive node classification task on citation datasets.
The values of the best performance are marked in bold. 
The baseline methods include the single GNN and vanilla KD method.
We observe that all the GNNs with the proposed OAD framework gain improvement over vanilla KD method and single GNNs. 
Specifically, on the Citeceer dataset, our model can bring 0.74\% and 2.05\% performance improvement for GCN and GraphSAGE, respectively. 
This demonstrates the effectiveness and the generalization of the proposed OAD framework.

\begin{table*}[htbp]
\label{citation architecture}
\centering
\caption{Network architecture used in citation networks.}
\begin{tabular}{llllll}
\hline
Dataset                   & Network           & Layers & Features & Attention heads & Parameters   \\
\hline
\multirow{9}{*}{Cora}     & GCN-Teacher       & 2      & 128,7     & /               &      0.18M        \\
                          & GCN-Student       & 2      & 16,7     & /            &   0.02M              \\
                          & GCN-Discriminator       & 1      & 1     & /            &   0.017K              \\
                          & GAT-Teacher       & 2      & 128,7     & 8,1             &    1.47M          \\
                          & GAT-Student       & 2      & 8,7      & 8,1             &      0.09M        \\
                         & GAT-Discriminator       & 2      & 16,1      & /             &      1.06K        \\
                          & GraphSAGE-Teacher & 2      & 128,7     & /               &        0.37M      \\
                          & GraphSAGE-Student & 2      & 16,1     & /               &   0.05M           \\
                          & GraphSAGE-Discriminator & 2      & 16,1     & /               &      0.29K        \\
\hline
\multirow{9}{*}{Citeceer} & GCN-Teacher       & 2      & 128,6     & /               &    0.47M       \\
                          & GCN-Student       & 2      & 16,6     & /               &      0.06M  \\
                          & GCN-Discriminator      & 1      & 1     & /               &      0.017K  \\
                          & GAT-Teacher       & 2      & 128,6     & 8,1             &      3.80M   \\
                          & GAT-Student       & 2      & 8,6      & 8,1             &   0.24M     \\
                          & GAT-Discriminator       & 2      &  16,1     & /             &   1.06K      \\
                          & GraphSAGE-Teacher & 2      & 128,6     & /               &      0.95M \\
                          & GraphSAGE-Student & 2      & 16,6     & /               &       0.12M \\
                          & GraphSAGE-Discriminator & 2      & 16,1     & /               &      0.29K        \\
\hline
\multirow{9}{*}{Pubmed}   & GCN-Teacher       & 2      & 128,3     & /               &         0.06M \\
                          & GCN-Student       & 2      & 16,3     & /               &       0.008M       \\
                          & GCN-Discriminator      & 1      & 1     & /               &      0.017K  \\
                          & GAT-Teacher       & 2      & 128,3     & 8,8             &     0.54M         \\
                          & GAT-Student       & 2      & 8,3      & 8,8             &       0.03M       \\
                         & GAT-Discriminator       & 2      &  16,1     & /             &   1.06K      \\
                          & GraphSAGE-Teacher & 2      & 128,3     & /               &    0.13M          \\
                          & GraphSAGE-Student & 2      & 16,3     & /               &     0.02M         \\
                         & GraphSAGE-Discriminator & 2      & 16,1     & /               &      0.29K        \\
\hline
\end{tabular}
\end{table*}

\begin{table}
  \caption{Top-1 Accuracy (\%) of different methods on three citation datasets. The results are obtained under the statistical significance level $\alpha<0.05$. The results of teacher and student ensemble are also included. The best results are in bold.}
  \label{citation result}
  \centering
  \begin{tabular}{cccc}
    \toprule
    Model     &  Cora   &    Pubmed   &   Citeceer \\
    \midrule
    GCN & 80.0 & 78.9 &  65.7 \\
    GCN+KD  &  80.4 &  79.0  &  66,5              \\
    GCN+OAD     & \textbf{80.9}   &  \textbf{79.3}  & 
    \textbf{67.5}  \\
    Improve \% &0.62\% &0.38\% &1.50\% \\
    \midrule
    GCN Ensemble    &  80.3  &  79.0  &   66.0   \\
    GCN Teacher  &  81.5  & 79.2 & 68.2   \\
    \midrule
    GAT     &  81.9   &  77.0    &  68.9   \\
    GAT+OAD     &  \textbf{82.2}  &  \textbf{78.5}  &
    \textbf{69.6}  \\
    GAT+KD  &  81.2 &  78.0 &  68.6 \\
    Improve \% &1.23\% &0.77\% &1.01\% \\
    \midrule
    GAT Ensemble &  82.1  &  77.5 &   69.9  \\
    GAT Teacher &  82.2  & 77.9 & 69.9   \\
    \midrule
    GraphSAGE    &  80.3   &  76.5       &  68.0   \\
    GraphSAGE+OAD   &  \textbf{81.6}   &  \textbf{77.4} &
    \textbf{69.7}    \\
    GraphSAGE+KD &  81.0  &  76.9  &  68.3   \\
    Improve \% &0.74\% &0.65\% &2.05\% \\
    \midrule
    GraphSAGE Ensemble  &  81.2  &  77.1  &   69.4  \\
    GraphSAGE Teacher &  81.7  & 78.3 & 69.9   \\
    \bottomrule
  \end{tabular}
\end{table}

\subsection{Inductive node classification on PPI dataset}
\label{subsection: ppi}
In this section, we evaluate the proposed OAD framework by inductive node classification task on Protein-to-Protein Interaction (PPI) dataset \citep{zitnik2017predicting}. 
PPI dataset contains 24 graphs corresponding to different human issues. 
The average number of nodes per graph is 2372.
Each node has 50 features that encode information about the positional gene set and immunological characters.
The total number of edges is 818716, constructed by prepossessing data provided by \citep{hamilton2017inductive}.
The dimension of the label for each node is 121. 
We follow the same data splitting protocol as \citep{velivckovic2018graph}, in which the number of graphs for training, validation and testing are 20, 2 and 2, respectively. 
Note that the testing graphs are completely undetectable during training. For PPI dataset, each node may belong to multiple categories and we evaluate the overall F1 score of the entire graph.

Table~\ref{ppi network architecture} summarizes the students' and teacher's network structures used in the experiment. The group size is set as 4. As can be seen from Table \ref{ppi network architecture}, \textit{the total parameters of student networks and extern discrminators are less than one teacher network}. For our model, students in the group warmup with cross entropy loss for 50 epochs, then they learn mutually for 50 epochs. For the single GNN and LSP \citep{yang2020distilling}, the total number of training epoch is 100. For all the comparisons, the learning rate, weight decay and optimizer are set as 0.005, 0 and Adam, respectively. The aggregation type of GraphSAGE is "mean". For our model, we adopt two-layer GCNs as discriminators in feature level distillation with 64 hidden neurons. The input of discriminator is the last hidden layers (the layer before last convolution layer) of GNNs. The loss coefficients is set as $\alpha=1, \beta=0.1$ for both GAT and GraphSAGE. All the results are generated by our implementations except LSP \citep{yang2020distilling} which we run the codes released by authors.
We also include the performance of the teacher model and an ensemble model. The group size of the ensemble model is same as OAD. Other hyperparameters settings of ensemble are the same as single GNN except group size.
Note that we evaluate the performance of all the student models and report the best case for our model. We repeat every experiment 4 times with different random seeds to initialize and report the average and variance.

Our baselines include some offline knowledge distillation methods: KD \citep{hinton2015distilling}, FitNet \citep{romero2014fitnets}, AT \citep{zagoruyko2016paying} and LSP \citep{yang2020distilling}. For these methods, we firstly train a common large enough teacher model as described in Table \ref{ppi network architecture}. Then, these KD methods will distill the knowledge of this teacher. The baselines also include an classic online knowledge distillation method - DML \citep{zhang2018deep}. The evaluation of DML is same as OAD by averaging the performance of all group members. We adopt GAT \citep{velivckovic2018graph} and GraphSAGE \citep{hamilton2017inductive} as the skeleton models. The student architecture of all the baseline methods are same.

We compare the F1 Score of different methods. The F1 score is defined as: 
\begin{equation}
    F1\_Score = \dfrac{2*P*R}{P+R}
\end{equation}
where $P$ is precision and $R$ is the recall. Table~\ref{ppi result} summarizes the results on the PPI dataset.
The values of the best performance are marked in bold.  
We observe that training with a pretrained teacher model will negatively influence the performance of GNN, while our proposed OAD can boost its performance.
Specifically, compared to single GNN model, proposed OAD method has 0.85\% and 2.27\% improvement for GAT and GraphSAGE, respectively. Moreover, OAD exceeds other vanilla knowledge distillation variants by more than 2\% for both GAT and GraphSAGE, indicating "virtual" teacher paradigm is more effective for GNN.
Compared to DML, proposed OAD method capture both local and global knowledge of GNN and improves by 0.35\% and 1.27\% for GAT and GraphSAGE, respecitively.

\begin{table*}
  \caption{Network architectures used on PPI dataset.}
  \label{ppi network architecture}
  \centering
  \begin{tabular}{lllll}
    \toprule
    Network    & Layers    & Features & Attention Heads & Parameters \\
    \midrule
    GAT-Teacher & 3  &  256,256,121 & 4,4,6 & 3.64M\\
    GAT-Student & 5  &  64,64,64,64,121 & 2,2,2,2,2 & 0.17M\\
    GAT-Discriminator &  2  &  64,1  &  /  &  0.008M                     \\
    GraphSAGE-Teacher & 3  & 256,256,121  & / & 0.22M \\
    GraphSAGE-Student & 5  & 64,64,64,64,121  & / & 0.047M \\
    GraphSAGE-Discriminator &  2  & 64,1 & / & 0.004M \\
    \bottomrule
  \end{tabular}
\end{table*}

\begin{table}
  \caption{F1 scores of different methods on PPI dataset. The results are obtained under the statistical significance level $\alpha<0.05$. The results of teacher and student ensemble are also included. The best results are in bold.}
  \label{ppi result}
  \centering
  \begin{tabular}{ccc}
    \toprule
    Method     & GAT            & GraphSAGE \\
    \midrule
    Student    & 90.13 $\pm$ 0.30   &  76.03 $\pm$ 0.61  \\
    Fitnet  & 87.63 $\pm$ 0.37  &  67.75 $\pm$ 0.96  \\
    AT  & 85.60 $\pm$ 1.20   &  74.45 $\pm$ 1.18 \\
    LSP  & 89.60 $\pm$ 0.62   &  74.83 $\pm$ 0.66  \\
    KD & 90.35 $\pm$ 0.33  & 76.28 $\pm$ 0.66 \\ 
    DML & 90.63 $\pm$ 0.50  & 77.03 $\pm$ 0.33  \\
    OAD  & \textbf{90.98 $\pm$ 0.46}  &  \textbf{78.30 $\pm$ 0.36} \\
    \midrule
    Teacher    & 97.0            &  86.3  \\
    Ensemble   & 94.8            &  80.1  \\
    \bottomrule
  \end{tabular}
\end{table}

\subsection{Object recognition on ModelNet40 dataset.}
\label{subsection:modelnet40}
ModelNet40 dataset is adopted for the object recognition task, consisting of 12311 meshed CAD models from 40 categories (9843 for training and 2468 for testing). 
Each CAD model belongs to one category, thus we use Top-1 and balanced accuracy as the evaluation metric.
Each model is constructed by a series of points with (x,y,z) coordinates. 
Following the experimental settings of \citep{qi2017pointnet}, we uniformly sample 1024 points for each model and then discard the original one. 
We use DGCNN as the skeleton model and construct the graph in the feature space by KNN algorithm as the previous work \citep{wang2019dynamic}. 
It is worth noting that the graph varies in different layers and different training stages. 
For a fair comparison, we use the same student architecture as that of LSP  \citep{yang2020distilling}. 

We adopt the same teacher and student architectures as \citep{yang2020distilling}, which are summarized in Table~\ref{modelnet network architecture}. We set the group size of our OAD model as 4. As can be seen from Table~\ref{modelnet network architecture}, \textit{the total parameters of student networks and extern discrminators are less than one teacher network}. For online knowledge distillation, the epoch of warmup learning and mutual learning are set as 100 and 150, respectively. For fair comparison, the single DGCNN is trained for 250 epochs. For all settings, we use SGD with momentum rate of 0.90 as optimizer. The learning rate, weight decay, dropout rate are 0.1 and 0.0001 and 0.5, respectively. The discriminators for proposed model are two-layer MLPs with batch normalization, which take the last hidden layer (the layer before output layer) of DGCNN as input. We set $\alpha=1, \beta=0.05$ as loss weights.  We cite the results of Fitnet, AT and LSP in the original paper \citep{yang2020distilling}. 
We also include the performance of the teacher model and an ensemble model. The group size of the ensemble model is same as OAD. Other hyperparameters settings of ensemble are the same as single GNN except group size.
Note that we evaluate the performance of all the student models and report the best case for our model. We repeat every experiment 4 times with different random seeds to initialize and report the average. 

We compare the recognition accuracy of different methods. Table~\ref{modelnet result} summarizes the results of object recognition on the ModelNet40 dataset. 
The values of the best performance are marked in bold. 
We can observe that the performance of our proposed model exceeds all the comparison methods. 
Besides, OAD can obtain a similar Top-1 accuracy score and a higher balanced accuracy score than the teacher model. 
This indicates that stacking several lightweight models with a significantly less number of parameters can perform almost as well as the teacher model. 
Another interesting observation is that most baseline methods perform worse than the student model, which is explainable as the graphs varies during training.
This further indicates the advantage of the OAD framework in adapting the changes of the graph. 

\begin{table*}
  \caption{Network architectures used on ModelNet40 dataset.}
  \label{modelnet network architecture}
  \centering
  \begin{tabular}{llllll}
    \toprule
    Network    & Layers    & Feature maps &  MLPs  &  K & Parameters \\
    \midrule
    Teacher &  5  &  64,64,128,256,1024 & 512,256 &  20  &  1.81M  \\
    Student &  4  &  32,32,64,128       & 256     &  10  &  0.1 M  \\
    Discriminator & 2 & 32,1  &  / & / & 4.25K \\
    \bottomrule
  \end{tabular}
\end{table*}

\begin{table}
\caption{Testing accuracy of different methods on ModelNet40 dataset. The results are obtained under the statistical significance level $\alpha<0.05$. The results of teacher and student ensemble are also included. The best results are in bold.}
  \label{modelnet result}
  \centering
  \begin{tabular}{ccc}
    \toprule
    Method    &     Top-1 Accuracy       & Balanced Accuracy  \\
    \midrule
    Student    & 92.3    & 88.8 \\
    Fitnet  & 91.1   &  87.9 \\
    AT  & 91.6   &  87.9 \\
    LSP  & 91.9   &  88.6 \\
    OAD   & \textbf{92.7}   &  \textbf{89.3}  \\
    \midrule
    Teacher    & 92.9        &   89.3  \\
    Ensemble   & 93.2        &  90.0  \\
    \bottomrule
  \end{tabular}
\end{table}

\subsection{Dynamic graph}
\label{dynamic graph}
\begin{figure}[htbp]
  \centering
  \includegraphics[width=\textwidth]{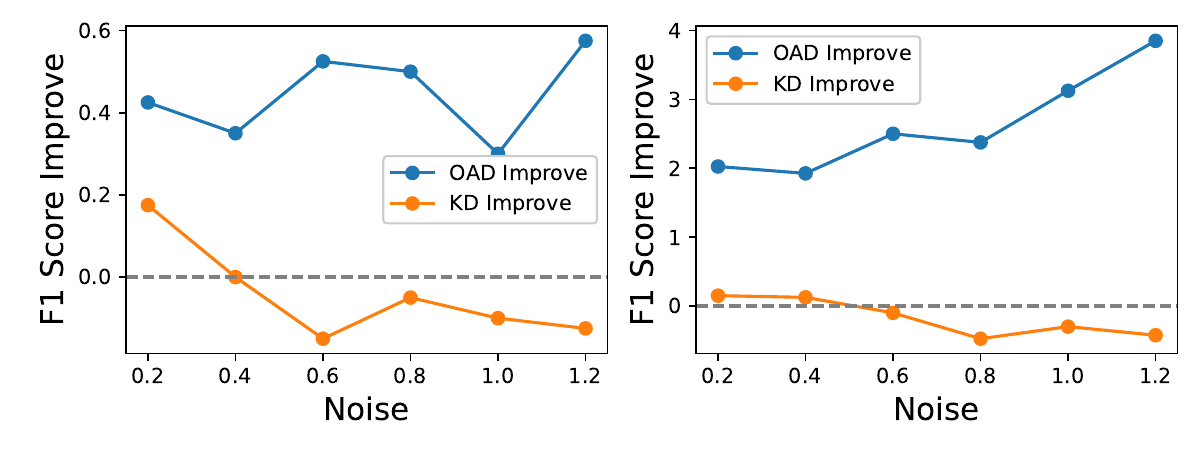}
  \caption{F1 score improvement w.r.t. a single GNN after adding ramdom noise on the node attributions of PPI dataset. The blue lines denote "OAD\_Improve" and the orange lines denote "KD\_Improve". The left subplot shows the experiment on GAT. The right subplot shows the experiment on GraphSAGE.}
  \label{fig:dynamic_noise}
\end{figure}

\begin{figure}[htbp]
  \centering
  \includegraphics[width=\textwidth]{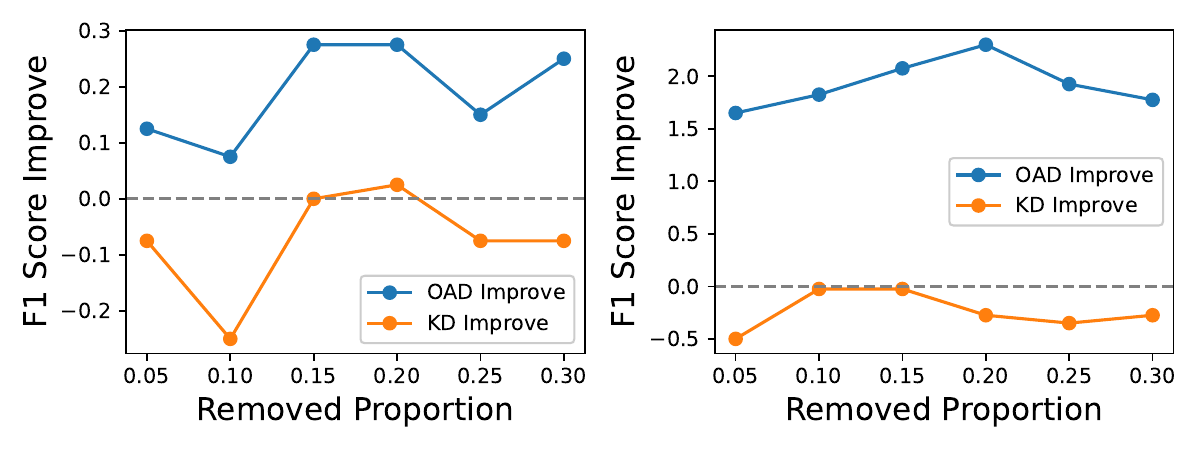}
\caption{F1 score improvement w.r.t. a single GNN after removing certain proportions of edges of the graphs of PPI dataset. The blue lines denote "OAD\_Improve" and the orange lines denote "KD\_Improve". The left subplot shows the experiment on GAT. The right subplot shows the experiment on GraphSAGE.}
  \label{fig:dynamic_remove}
\end{figure}

In this section, we compare the performance of a single GNN, conventional knowledge distillation and our proposed OAD model on dynamic graphs. 
The changes of a graph may be caused by node attributes variation or graph structure evolution. Corresponding to above two circumstances, we conduct two simulation experiments on PPI dataset.
Firstly, we add random noises on each node's attribution (This experiment is denoted as "Dynamic-A"). The noise has zero means and frequencies ranging from 0.2 to 1.2. 
Then, we randomly remove certain proportions of edges on the graph (This experiment is denoted as "Dynamic-B"). The removed proportion ranges from 0.05 to 0.3. 
Consider that GNNs may fail to train after its edges being removed, we add self loops on the isolated nodes. We adopt GAT and GraphSAGE as the skeleton networks. 
Their architectures are summarized in Table \ref{ppi network architecture}. For knowledge distillation, we pretrain a much larger teacher GNN on original graph and then transfer its knowledge about logits distribution to a student GNN trained on dynamic graphs. 
For OAD and the single GNN, the training process is completely on dynamic graphs. 
Other experimental details are the same as Subsection \ref{subsection: ppi}. We compute the improvement of KD compared to the single student GNN (denoted as "KD\_Improve") and improvement of OAD compared to the single student GNN (denoted as "OAD\_Improve") as follows:

\begin{equation}
\begin{split}
    KD\_Improve = F1\_Score(KD) - F1\_Score(single)  \\
    OAD\_Improve = F1\_Score(OAD) - F1\_Score(single)
\end{split}
\end{equation}
where $F1\_Score(single), F1\_Score(KD), F1\_Score(OAD)$ are F1 scores of single GNN, KD and OAD, respectively. 
The results of "Dynamic-A" and "Dynamic-B" are presented in Figure \ref{fig:dynamic_noise} and \ref{fig:dynamic_remove}, respectively. From Figure \ref{fig:dynamic_noise}, we observe that as the graph changes more drastically, the value of "KD Improve" drops. 
When the frequency of random noise exceeds 0.6, the knowledge transferred from the static teacher is even harmful to the student. 
On the other hand, our proposed OAD model exceeds student GNN on all dynamic graphs, especially when the dynamic graph is greatly different from the original graph. 
Note that there are no explicit relationships between "F1 score improvement" and noise rate.
The value of "F1 Score Improve" fluctuates with the increase of noise rate, as OAD model and student GNN may have different parameter initializations on different noise rate and edge removal proportion settings.
From Figure \ref{fig:dynamic_remove}, we observe that for both GAT and GraphSAGE, the values of "KD Improve" are marginally around 0, indicating the knowledge obtained from the old graph is not effective on the new graph. Meanwhile, OAD maintains the superiority on the dynamic graphs. 
The reason behind this phenomenon may be that all students are trained on the graph of same dynamic frequency simultaneously in our model, they can capture different characteristics of the dynamic graph and knowledge consistency is retained.
 
\begin{figure*}[htbp]
  \centering
  \includegraphics[width=\textwidth]{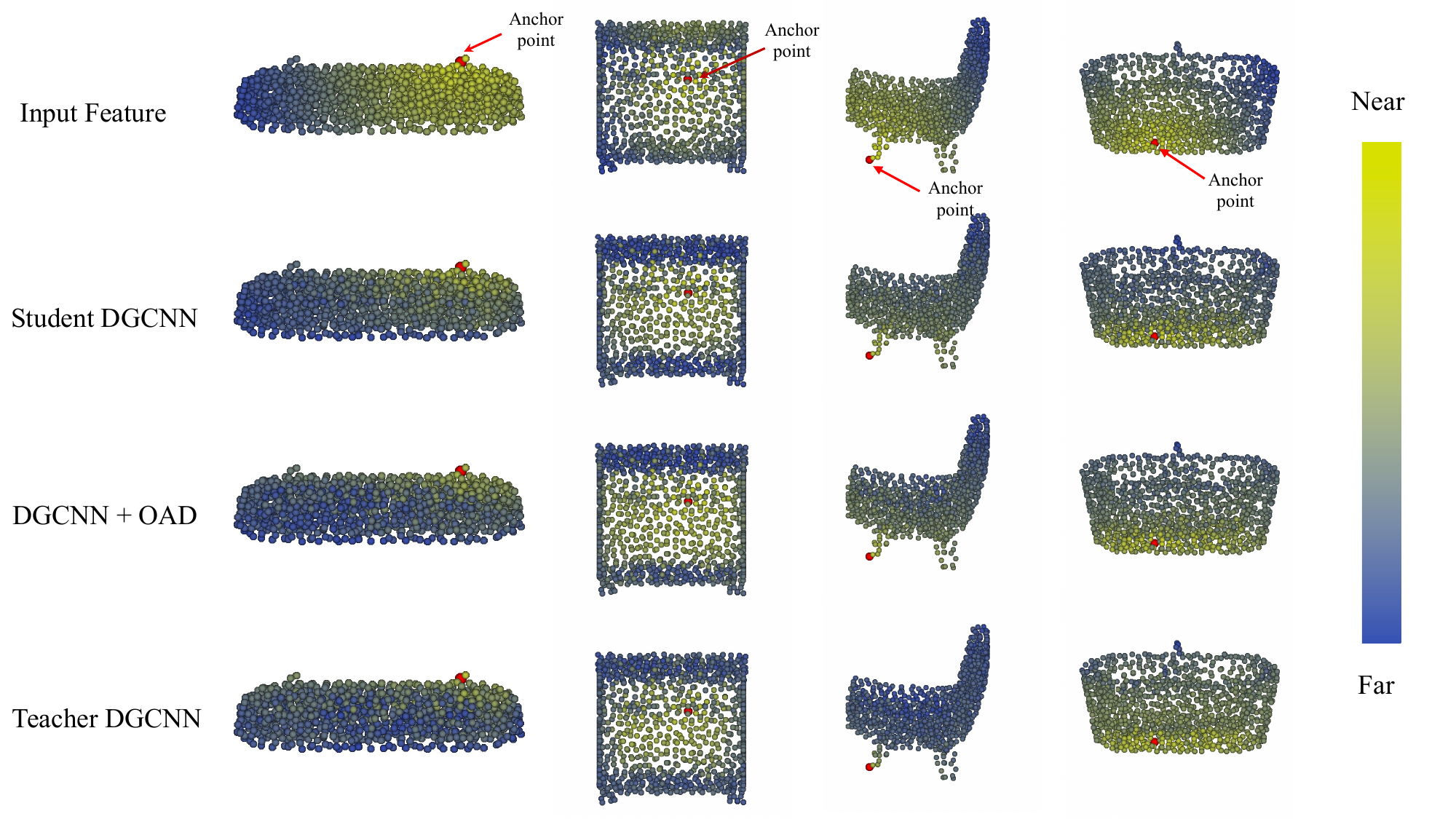}
  \caption{Visualization of features learned by different models, deprecated by the distance between anchor points (colored in red) and other points. From left to right: original input features; features learned by student DGCNN; features learned by DGCNN with OAD framework and features learned learned by teacher DGCNN. Best viewed in color.}
  \label{fig:visualization}
\end{figure*}

\subsection{Benefit of mutual learning}
\label{subsection:benefit of mutual learning}
\begin{figure}
  \centering
  \includegraphics[width=\textwidth]{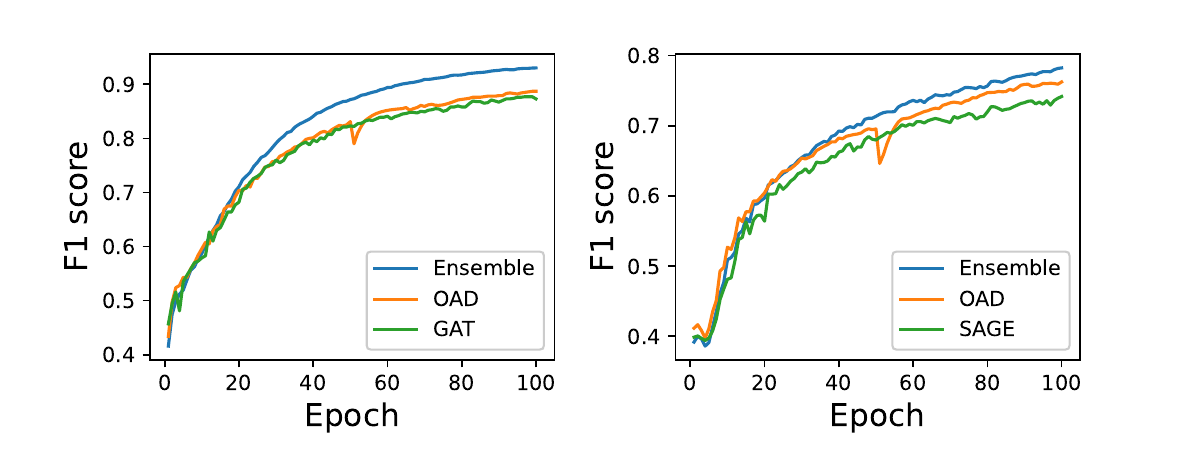}
  \caption{The training F1 scores of Ensemble, OAD and a vanilla GNN on PPI dataset.}
  \label{fig:analysis}
\end{figure}

In this section, we further analyze the benefit of group-based mutual learning. 
In Figure \ref{fig:analysis}, we compare the validation micro F1 scores of ``ensemble'', OAD and a vanilla GNN on the PPI dataset during the training phase. 
The experimental setup is the same as Subsection 4.2. 
We plot the validation F1 scores from epoch 0 to epoch 100. Our OAD model is warmed up from epoch 0 to 50 without group knowledge transfer. 
From 50 to 100 epoch, additional discriminators are introduced for mutual learning. It can be seen that the F1 scores of ``ensemble'' are consistently higher than the GNN trained with OAD, which indicates that the generated group knowledge may act like a ``virtual" teacher to help improve the performance of the GNN model. We observe that the performance drops slightly at around epoch 50, because the parameters of newly attached discriminators are randomly initialized. After several epochs, as the discriminators learn the difference of GNN students' features, they can aid the training process. 

\subsection{Ablation Study}
    To further evaluate the effectiveness of each component in our OAD framework, we perform an ablation study on the PPI dataset and ModelNet40 dataset. The results are presented in Table~\ref{Ablation study on PPI dataset} and \ref{Ablation study on modelnet dataset}. We analyze the contributions of each component as follows:
    
    (1) w/o $\mathcal{L}_{B}$ refers to removing the global knowledge distillation module ($\mathcal{L}_{B}$) of the OAD framework. On the PPI dataset, the ablation of this module decreases the micro-F1 scores of GAT and GraphSAGE by 0.4\% and 1.2\%, respectively. As for the ModelNet40 dataset, DGCNN obtains 0.3\% lower top-1 accuracy and 0.2\% lower balanced accuracy without this module. This demonstrates that global knowledge distillation can bring benefits to the proposed framework. 
        
    (2) w/o $\mathcal{L}_{G}$ refers to removing the local knowledge distillation module ($\mathcal{L}_{G}$) of the OAD framework. As a result, all the external discriminators are removed and knowledge transfer on node embedding space is prohibited. This setting decreases the performance of GAT by 0.18\% and GraphSAGE by 1.22\%, respectively. On the ModelNet40 dataset, after removing local knowledge distillation module, Top-1 accuracy and balanced accuracy of DGCNN drops by 0.3\%. Thus, including the local knowledge distillation module can improve the performance of the OAD framework.

\begin{table}
  \caption{Ablation studies on PPI dataset.}
  \label{Ablation study on PPI dataset}
  \centering
  \begin{tabular}{ccc}
    \toprule
    Method     & GAT            & GraphSAGE \\
    \midrule
    w/o $\mathcal{L}_{B}$ &  90.58 $\pm$ 0.35  &  77.10 $\pm$ 0.56\\
    w/o $\mathcal{L}_{G}$ &  90.80 $\pm$ 0.34  &  77.08 $\pm$ 0.54  \\
    OAD & 90.98 $\pm$ 0.46  &  78.30 $\pm$ 0.36 \\
    \bottomrule
  \end{tabular}
\end{table}

\begin{table}
  \caption{Ablation studies on ModelNet40 dataset.}
  \label{Ablation study on modelnet dataset}
  \centering
  \begin{tabular}{ccc}
    \toprule
    Method    &     Top-1 Accuracy       & Balanced Accuracy  \\
    \midrule
    w/o $\mathcal{L}_{B}$  &  92.4   &  89.0   \\
    w/o $\mathcal{L}_{G}$ &  92.2   &  88.7     \\
    OAD                       & 92.7   &  89.3 \\
    \bottomrule
  \end{tabular}
\end{table}


\begin{table}
  \caption{Processing time of different methods on PPI dataset (seconds).}
  \label{processing time on ppi dataset}
  \centering
  \begin{tabular}{ccc}
    \toprule
         &     GAT       &  GraphSAGE  \\
    \midrule
    Teacher  &  679.8   &  610.4   \\
    Student  &  154.5   &  126.1     \\
    OAD ($M=2$) & 320.9  &  285.2 \\
    OAD ($M=4$)  &   526.1  &  468.0  \\
    OAD ($M=6$) &  894.4  &  825.5  \\
    OAD ($M=8$)  &  1153.2  &  1061.8  \\
    \bottomrule
  \end{tabular}
\end{table}

\subsection{Comparisons of processing time}
    Compared to vanilla KD approaches, a pretrained teacher GNN model is not required by OAD method. Thus, OAD can reduce the computation load and save the processing time. To verify this point, we compare the training time of KD and OAD on PPI dataset using the same experiment settings as Section \ref{subsection: ppi}. The time of data loading and evaluating models is excluded. The results are shown in Table \ref{processing time on ppi dataset}. Note that the training time of KD methods is the sum of teacher's training time and student's training time. It is shown in Table \ref{processing time on ppi dataset} that the processing time of OAD is approximately 2/3 of the processing time of KD. 
    
\begin{figure}
    \centering
    \includegraphics[width=\textwidth]{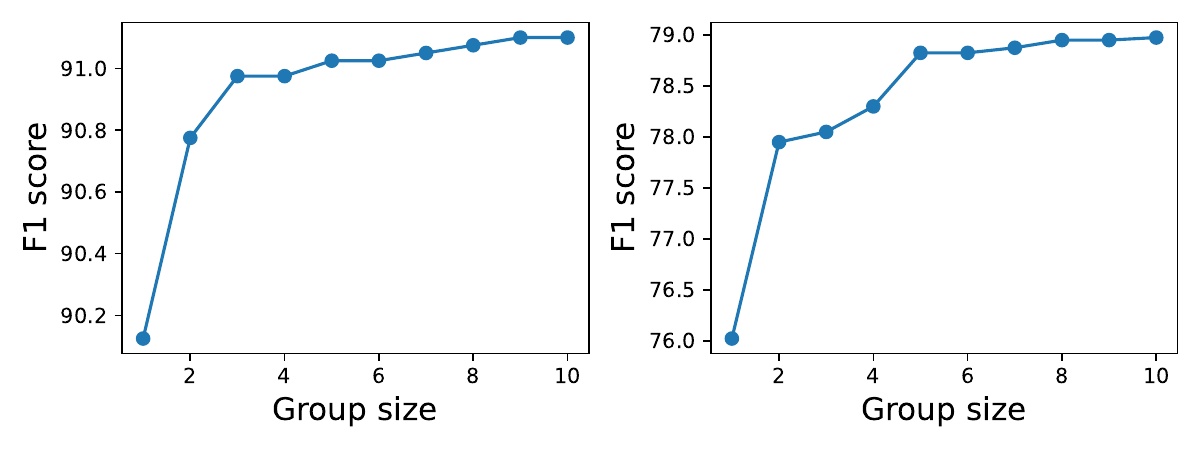}
    \caption{The F1 scores on PPI dataset with different group size. Left: GAT. Right: GraphSAGE.}
    \label{fig:groupsize}
\end{figure}

\begin{table}
  \caption{F1 score of GraphSAGE on PPI dataset w.r.t. various temperatures.}
  \label{various temperature}
  \centering
  \begin{tabular}{ccccccc}
    \toprule
    $T$     &  1.0 & 2.0 & 3.0 & 4.0 & 5.0 & 6.0   \\
    \midrule
    OAD  & 78.30  &  78.63  &  78.30  &  78.87  &  78.83  &  78.67 \\
    KD & 76.67 & 76.50 & 76.28 & 76.03 & 75.97 & 76.43 \\
    \bottomrule
  \end{tabular}
\end{table}

\begin{table}
  \caption{F1 score of GraphSAGE on PPI dataset w.r.t. various loss weights.}
  \label{various loss weight}
  \centering
  \begin{tabular}{ccccccc}
    \toprule
    $\alpha$ &  0.5 & 1.0 & 1.5 & 2.0 & 2.5 & 3.0   \\
    \midrule
    OAD & 78.37 & 78.30 & 78.70 & 78.20 & 78.57 & 78.47 \\
    \midrule
    $\beta$  &  0.05 & 0.10 & 0.15 & 0.20 & 0.25 & 0.30   \\
    \midrule
    OAD & 78.57 & 78.30 & 78.70 & 78.20 & 78.57 & 78.47 \\
    \bottomrule
  \end{tabular}
\end{table}

\subsection{Sensitivity Analysis}
\paragraph{Group size} We evaluate the impact of group size (number of student GNN models) on the performance of our proposed OAD framework. Figure \ref{fig:groupsize} illustrates the performance concerning the group size ranging from 1 to 10. We observe that the F1 scores of GAT and GraphSAGE will generally increase as the student group is enlarged. We also observe that when the group size exceeds a specific threshold, the benefit becomes gradually marginal until convergence due to the capacity saturation.
\paragraph{Temperature} We then evaluate the influence of temperature $T$ to our OAD method. We conduct the experiments on PPI dataset using GraphSAGE as the student model. The hyperparameters setting is same as Section \ref{subsection: ppi} except temperature used in global knowledge distillation. The results are shown in Table \ref{various temperature}. We observe that the F1 scores of OAD are substantially higher than vanilla KD method in all temperature settings. Thus, we conclude that proposed OAD method is not sensitive to the temperature chosen.
\paragraph{Loss coefficients} We finally evaluate the influence of loss coefficients $\alpha$ and $\beta$. We testify the F1 score of GraphSAGE model on PPI dataset. The hyperparameters setting is same as Section \ref{subsection: ppi} except $\alpha$ and $\beta$ settings. Note that $\alpha=1$ is fixed when varying $\beta$ and $\beta=0.1$ is fixed when varying $\alpha$. The results are shown in Table \ref{various loss weight} We observe that the performance of OAD slightly fluctuates with different loss weights. Thus OAD is robust to loss weight selection.

\subsection{Visualization}
In this section, we visualize the features learned by different student GNN  models.
We extract the last layer of the student GNN models as features and visualize the distance between a randomly selected anchor point (colored in red) and other points on the model. 
Figure \ref{fig:visualization} illustrates the Euclidean distance of points in the input space, from top to bottom are feature spaces of single student DGCNN model, DGCNN with OAD framework and the teacher DGCNN model. From left to right are five different samples in ModelNet40 dataset. We use the feature space of teacher DGCNN model as the reference. We can observe that the feature structure of the proposed model is similar to the feature structure of the teacher. This demonstrates the capability of the representations learned by the OAD framework.

\section{Conclusions and Future Works}
\label{conclusion}
Knowledge distillation has become a promising technique to improve the performance of CNNs, under the assumption that teacher and student models are trained on the identical data distribution.
However, since the topological structure and node attributes of graph data are likely to evolve, this fundamental assumptions of knowledge distillation may not hold for Graph Neural Networks (GNNs), leading to the sub-optimal solution.
In this paper, we propose online knowledge distillation for graph neural networks to tackle this challenge.
More specifically, we train a group of student GNN models simultaneously and with the guidance of a virtual dynamic teacher, the performance of each member is improved by distilling both the local and global knowledge. 
Adversarial cyclic learning is utilized to effectively and efficiently exploit the complicated information contained in the graph topology and node attributes. Extensive experiments on citation dataset, PPI dataset and ModelNet40 dataset demonstrate the effectiveness of the proposed OAD framework. 

\paragraph{Future Works} There are some possible future directions of this work. Firstly, Graph Transformer models have shown great potential on graph learning. Compared with GNN, Graph Transformers are heavier models, thus applying knowledge distillation on Graph Transformer is more meaningful than GNN. Therefore, we will study how to distill the knowledge of Graph Transformer. Secondly, although online knowledge distillation may alleviate the problem of distribution shift on graph data, its processing speed is slower as the group size increases. How to maintain the performance of OAD while making it more efficient should be studied in the future work. 

\section*{Acknowledgments}
This work is supported by the Starry Night Science Fund of Zhejiang University Shanghai Institute for Advanced Study (Grant No: SN-ZJU-SIAS-001), National Natural Science Foundation of China (Grant No: U1866602), National Natural Science Foundation of China (Grant No: 621062219), Zhejiang Provincial Natural Science Foundation of China under \\ Grant No. LTGG23F030005, Ningbo Natural Science Foundation (Grant No: 2022J183) and the advanced computing resources provided by the Supercomputing Center of Hangzhou City University.

\bibliographystyle{elsarticle-harv} 
\bibliography{ref2}

\end{document}